\documentclass[journal,onecolumn]{IEEEtran}
\ifCLASSINFOpdf
\else
\fi
\usepackage{epsfig}
\usepackage{graphicx}
\usepackage{amsmath}
\usepackage{amssymb}
\usepackage{commath}
\usepackage{bbding}
\usepackage{cite}
\usepackage{cleveref}
\usepackage{booktabs,multirow}

\usepackage{epstopdf}

\usepackage{amsmath,graphicx}
\usepackage{amsfonts}
\usepackage{subfigure}

\usepackage{bbm,booktabs,tabulary}

\begin{document}
%
\title{Single Image Action Recognition by Predicting Space-Time Saliency}
%
%
%

\author{Marjaneh Safaei and Hassan Foroosh
\thanks{Marjaneh Safaei and Hassan Foroosh are with the Department of Computer Science, University of Central Florida, Orlando,
FL, 32816 USA (e-mail: msafaei@cs.ucf.edu, foroosh@cs.ucf.edu).}
}

\maketitle

\begin{abstract}
We propose a novel approach based on deep Convolutional Neural Networks (CNN) to recognize human actions in still images by predicting the future motion, and detecting the shape and location of the salient parts of the image. We make the following major contributions to this important area of research: (i) We use the predicted future motion in the static image (Walker et al., 2015) as a means of compensating for the missing temporal information, while using the saliency map to represent the the spatial information in the form of location and shape of what is predicted as significant. (ii) We cast action classification in static images as a domain adaptation problem by transfer learning. We first map the input static image to a new domain that we refer to as the Predicted Optical Flow-Saliency Map domain (POF-SM), and then fine-tune the layers of a deep CNN model trained on classifying the ImageNet dataset to perform action classification in the POF-SM domain. (iii) We tested our method on the popular Willow dataset. But unlike existing methods, we also tested on a more realistic and challenging dataset of over 2M still images that we collected and labeled by taking random frames from the UCF-101 video dataset. We call our dataset the UCF Still Image dataset or UCFSI-101 in short. Our results outperform the state of the art.
\end{abstract}



\section{Introduction}
\label{sec:intro}

Human action recognition from visual data has a wide range of applications in areas such as image annotation \cite{Tariq_etal_2017_2,Tariq_etal_2017,tariq2013exploiting,tariq2014scene}, surveillance and image retrieval \cite{Junejo_etal_2007,Junejo_Foroosh_2008,Sun_etal_2012,junejo2007trajectory,sun2011motion,Ashraf_etal2012,sun2014feature,Junejo_Foroosh2007-1,Junejo_Foroosh2007-2,Junejo_Foroosh2007-3,Junejo_Foroosh2006-1,Junejo_Foroosh2006-2,ashraf2012motion,ashraf2015motion,sun2014should},,  and video post-production and editing \cite{Cao_etal_2005,Cao_etal_2009,shen2006video,balci2006real,xiao20063d,moore2008learning,alnasser2006image,Alnasser_Foroosh_rend2006,fu2004expression,balci2006image,xiao2006new,cao2006synthesizing}. 
 to name a few.


The literature on human action recognition is primarily dominated by methods that rely on video data as input \cite{Shen_Foroosh_2009,Ashraf_etal_2014,Ashraf_etal_2013,Sun_etal_2015,shen2008view,sun2011action,ashraf2014view,shen2008action,shen2008view-2,ashraf2013view,ashraf2010view,boyraz122014action,Shen_Foroosh_FR2008,Shen_Foroosh_pose2008,ashraf2012human}. 
Video data has the advantage of providing temporal information, which plays an important role in distinguishing different actions. There has also been methods that attempt to tackle viewpoint variations \cite{Shen_Foroosh_2009,Ashraf_etal_2014,Ashraf_etal_2013,shen2008view,ashraf2014view,shen2008view-2,ashraf2013view,ashraf2010view,Shen_Foroosh_FR2008,ashraf2012human}, although such methods may require calibration across view \cite{Cao_Foroosh_2007,Cao_Foroosh_2006,Cao_etal_2006,Junejo_etal_2011,cao2004camera,cao2004simple,caometrology,junejo2006dissecting,junejo2007robust,cao2006self,foroosh2005self,junejo2006robust,Junejo_Foroosh_calib2008,Junejo_Foroosh_PTZ2008,Junejo_Foroosh_SolCalib2008,Ashraf_Foroosh_2008,Junejo_Foroosh_Givens2008,Lu_Foroosh2006,Balci_Foroosh_metro2005,Cao_Foroosh_calib2004,Cao_Foroosh_calib2004,cao2006camera}. Action recognition in still images is, on the other hand, more challenging due to the absence of motion information. The problem becomes even more challenging when there is no contextual information, e.g. when no objects (other than the human) are available in the image. Interestingly, humans perform this task almost effortlessly. Humans have a remarkable ability to recognize an action in a still image, despite the lack of temporal information. Human brain is able to not only recognize what is present in the image but also predict what action may take place next. Therefore, predicting the future motion plays a very important role in the prediction of a human action, especially when the action relies mainly on human body motions, and not the human-object interactions.

Given the tremendous growth in the number of images on the Web, it is of paramount importance to automate the analysis of human action in still images. Remarkable progress has been made in human action recognition in video data. In contrast, action recognition in still images remains more challenging and less attended by researchers. The problem may be sometimes exacerbated by the quality of the image, in which case preprocessing steps can be helpful \cite{Foroosh_2000,Foroosh_Chellappa_1999,Foroosh_etal_1996,Cao_etal_2015,berthod1994reconstruction,shekarforoush19953d,lorette1997super,shekarforoush1998multi,shekarforoush1996super,shekarforoush1995sub,shekarforoush1999conditioning,shekarforoush1998adaptive,berthod1994refining,shekarforoush1998denoising,bhutta2006blind,jain2008super,shekarforoush2000noise,shekarforoush1999super,shekarforoush1998blind}, also actions may be located in different parts of the image or present at different visual scales, in which case global registration methods may be helpful \cite{Foroosh_etal_2002,Foroosh_2005,Balci_Foroosh_2006,Balci_Foroosh_2006_2,Alnasser_Foroosh_2008,Atalay_Foroosh_2017,Atalay_Foroosh_2017-2,shekarforoush1996subpixel,foroosh2004sub,shekarforoush1995subpixel,balci2005inferring,balci2005estimating,foroosh2003motion,Balci_Foroosh_phase2005,Foroosh_Balci_2004,foroosh2001closed,shekarforoush2000multifractal,balci2006subpixel,balci2006alignment,foroosh2004adaptive,foroosh2003adaptive}. However, in general, the challenge mainly resides in the lack of temporal information, as further described below.

\section{Related Work}
\label{sec:related}
Human action recognition in still images has gained increasing attention in recent years (Zhang et al., 2016; Oquab et al., 2014; Gkioxari et al., 2014; Khan et al., 2013) due to its challenging nature and its importance in applications such as image search and retrieval. 
A comprehensive survey was performed by Guo and Lai, where existing action recognition methods are categorized based on low-level features (e.g SIFT \cite{lowe2004distinctive} and HOG \cite{dalal2005histograms}) and high-level cues, such as attributes \cite{bourdev2011describing,yao2011human}, body parts and pose \cite{delaitre2010recognizing,zheng2012action, maji2011action}, which are challenging due to the limited number of poses they can detect and also the fact that many different human actions share almost the same poses, and human-object interaction \cite{prest2012weakly,yao2011human,yao2012recognizing}. Other approaches to action recognition employ Bag of Words (BOW)-based image representations \cite{khan2013coloring,khan2014semantic,sharma2012discriminative}.

Action recognition in still images has recently benefited from CNN models \cite{gkioxari2014r,hoai122014regularized,oquab2014learning}, which offer an outstanding performance. The tradeoff, however, is that training the CNNs requires millions of parameters and often a huge number of annotated images. This poses a challenge when fewer training data are available. CNNs are high-capacity classifiers with very large numbers of parameters that must be learned from training examples. Action recognition in still images suffers from lack of annotated images for a wide range of action classes. Recent works on single image action recognition focus on a limited number of action classes and primarily rely on human-object interaction, which requires tackling a very challenging problem of recognizing shapes and objects other than humans in the scene \cite{Shu_etal_2016,Milikan_etal_2017,Millikan_etal2015,shekarforoush2000multi,millikan2015initialized,Cakmakci_etal_2008,Cakmakci_etal_2008_2,Zhang_etal_2015,Lotfian_Foroosh_2017,Morley_Foroosh2017,Ali-Foroosh2016,Ali-Foroosh2015,Einsele_Foroosh_2015,ali2016character,Cakmakci_etal2008,damkjer2014mesh}. This poses a challenge when the action merely involves a human with no object interaction, or when the objects next to humans cannot be readily recognized. In such cases, the most salient parts of the body and their predicted future motion can play a crucial role in action recognition. 

It is worth mentioning that most of the large scale image datasets such as Caltech \cite{fei2006one}, Pascal VOC \cite{everingham2010pascal} and ImageNet \cite{imagenet_cvpr09} have been created for the purpose of object recognition, but not for action recognition. To address this problem, we have cast our problem as a domain-adaptation problem. We built a POF-SM dataset consisting of over \textit{2M} images. First, we sampled frames from each video in UCF-101 \cite{soomro2012ucf101} and collected a large dataset of still images consisting of action-labeled images. Then, we mapped all the images in this dataset into the new domain; i.e. POF-SM. The Predicted Optical Flow (POF) for each pixel in the image \cite{walker2015dense}, combined with the static Saliency Map (SM), yields a three-channel input data for our CNN, hereafter referred to as  $I\textsubscript{POF\textsubscript{h} POF\textsubscript{v}SM }$.  

\section{OUR APPROACH}
\label{sec:approach}

In this section, we first describe the proposed domain mapping to transfer data from the source domain of static raw images to the POF-SM domain. We then describe our deep CNN architecture, implementing the proposed domain adaptation, to transfer learning from raw image classification to still image action classification in the POF-SM domain. 

\subsection{Domain Mapping}
\label{ssec:domain-mapping}

As depicted in Figure 1, as part of our domain adaptation scheme, we convert the input images to $I\textsubscript{POF\textsubscript{h} POF\textsubscript{v}SM }$, i.e. a three-channel input data for our network. We use a CNN model similar to the one proposed by Walker et al. \cite{walker2015dense} to predict a dense optical flow. This optical flow represents how and where each pixel in the input static image is predicted to move. For this purpose, the optical flow vectors are first quantized into 40 clusters by \textit{k}-means. The problem is then treated in a manner similar to semantic segmentation, where each region in the image is classified as a particular cluster of the optical flow. A softmax loss layer at the output is then used for computing gradients. The softmax loss is spatial, summing over all the individual region losses. This leads to a $M\times N \times C$ softmax layer, where $M$, $N$ and $C$ represent the number of rows, columns and clusters, respectively. 

\begin{figure*}[h]

  \centering
  \centerline{\includegraphics[width=\textwidth]{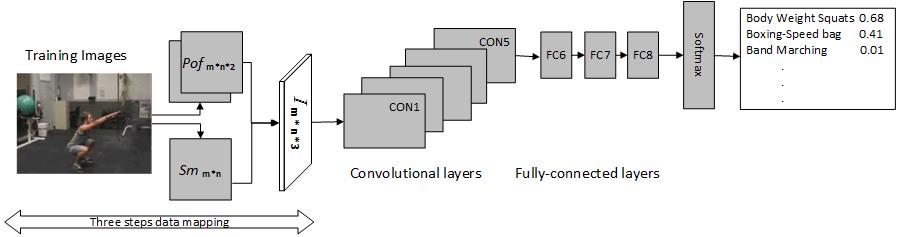}}
%
%
\caption{Schematic overview of our approach. We first map the input image \textit{I} onto $I\textsubscript{POF\textsubscript{h} POF\textsubscript{v}SM }$. We then employ deep features extracted from activation of the fully connected layers of the CNN and form the final prediction using the softmax operator.}
\label{fig:res}
\end{figure*}

\begin{figure}[h]

  \centering
  \centerline{\includegraphics[width=15.5cm]{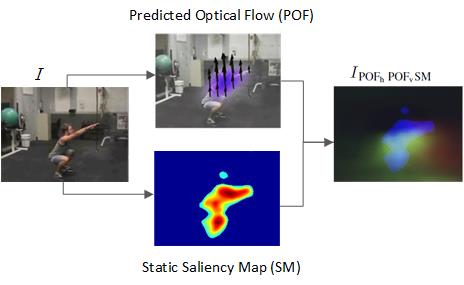}}

\caption{Example of mapping \textit{I} onto $I\textsubscript{POF\textsubscript{h} POF\textsubscript{v}SM }$  }
\label{fig:res}
%
\end{figure}


Let \textit{I} represent the image and \textit{Y} be the ground truth optical flow labels represented as quantized clusters. Then the spatial loss function ${L(I,Y)}$ as defined by Walker et al. is given by:
\begin{equation} \label{eq:1}
L(I,Y)=-\sum\limits_{i=1}^{M\times N}\sum\limits_{r=1}^C \mathbbm{1}(Y_i = r )\log F_{i,r}(I),
\end{equation} 
where $F\textsubscript{i,r}(I)$ represents the probability that the \textit{i}\textsubscript{th} pixel will move according to cluster \textit{r}, and $\mathbbm{1}(Y\textsubscript{i} = r )$ is an indicator function. A problem with this loss function is that it implicitly assumes a uniform probability mass function (pmf) for the motion clusters, which is very unlikely and prone to noise. Therefore, we modified Eq. \eqref{eq:1} in order to minimize the noise by taking into account only the $K$ most-likely clusters, i.e. the $K$ clusters with the highest probability. This amounts to replacing the second summation in Eq. \eqref{eq:1} with an order statistic filter as follows:

\begin{equation} \label{eq:2}
\hat{L}(I,Y)=-\sum\limits_{i=1}^{M\times N}\sum\limits_{r=1}^C \omega_r P_{i,(r)},
\end{equation} 
where $\omega_r$ are some weight factors, and 
\begin{equation}
P_{i,(r)}=\mathbbm{1}(Y_i = (r) )\log F_{i,(r)}(I)
\end{equation} 
is the pmf in descending order of values, i.e. $P_{i,(1)} \geq P_{i,(2)} \geq ... \geq P_{i,(C)}$. In our implementation, we set $K=10$ and assumed $\omega_r=\frac{1}{K}$ for $P_{i,(1)},\dots, P_{i,(K)}$, and $\omega_r=0$, otherwise. Essentially, we end up averaging over the probabilities of the $K$ most-likely clusters.

The SM channel of the target domain represents the static saliency map of the image using a bottom-up approach \cite{seo2009static}, where each pixel indicates the likelihood of saliency of a feature matrix given its surrounding feature matrices. We set all the values below some threshold $\tau$ in the SM channel to zero, in order to make sure to have a better localization and representation of the shape of the salient parts of the image. The threshold $\tau$ was selected by Otsu's method \cite{Otsu}.  

By generating the POF (horizontal and vertical components) and the saliency map for each image, we map the RGB images in the datasets to the target domain $I\textsubscript{POF\textsubscript{h} POF\textsubscript{v}SM }$. 

\subsection{Network Architecture}
\label{ssec:net-arch}

Our CNN network is similar to the standard seven-layer architecture proposed in \cite{krizhevsky2012imagenet}. The CNN architecture of \cite{krizhevsky2012imagenet} contains more that 60 million parameters and 650,000 neurons. This network is formed by five successive convolutional layers \textit{C1...C5} followed by three fully connected layers \textit{FC6...FC8}.
The three fully connected layers then compute $Y\textsubscript{6}= \sigma (W\textsubscript{6}Y\textsubscript{5}+ B \textsubscript{6})$, $Y\textsubscript{7}= \sigma (W\textsubscript{7}Y\textsubscript{6}+ B \textsubscript{7})$, $Y\textsubscript{8}= \psi (W\textsubscript{8}Y\textsubscript{7}+ B \textsubscript{8})$, where $Y\textsubscript{m}$ denotes the output of the \textit{m}\textsubscript{th} layer and $W\textsubscript{m}$ and $B\textsubscript{m}$ are trainable parameters of the \textit{m}\textsubscript{th} layer. Also, $\sigma(X)[i]=max(0,X[i])$ and $\psi(X)[i]=e^{X[i]}/\Sigma_j e^{X[j]}$ are the "ReLU" and "SoftMax" non-linear activation functions.

To accomplish transfer-learning by adapting to the new target domain of the input data, we changed the \textit{FC8} layer in order to adapt it to our target domain classes. In all our experiments, we keep some of the earlier layers fixed, and fine-tune some higher-level portions of the network, including \textit{C4}...\textit{FC7}, and finally train the new classifier layer \textit{FC8} from scratch. Even though it appears that we can afford to train a network from scratch when the target dataset is large enough, in practice it is quite often still beneficial to initialize the weights from a pre-trained model. In this case, we have enough data and confidence to fine-tune through the entire network. The new \textit{FC8} is trained on the target dataset from scratch with a higher learning rate. Our goal is to learn a mapping between the predicted optical flows combined with spatial information of the most salient parts of the image, and the action being performed in the image. 

Below, we denote the convolutional layers as CON(\textit{k},\textit{s}), which indicates that there are \textit{k} kernels, of size $s\times s$. We also denote the local response normalization layer as LRN, and the max-pooling layer as MP. All the strides are set to 1 except for the first layer, which is set to 4 during convolution. The stride for pooling is 2, and we set the pooling kernel size as $3\times 3$. Finally, FC(\textit{n}) denotes a fully connected layer with \textit{n} neurons. Our network architecture can be described as:
CON(96, 11)$\,\to\,$LRN$\,\to\,$MP$\,\to\,$CON(256,5)$\,\to\,$LRN$\,\to\,$MP$\,\to\,$CON(384,3)$\,\to\,$CON(384,3)$\,\to\,$CON(256,3)$\,\to\,$RLN$\,\to\,$MP$\,\to\,$FC(4096)$\,\to\,$FC(4096). We use a base learning rate of 0.001 and a step size of 70000 iterations. We use a smaller learning rate for layer weights that are being fine-tuned, in comparison to the weights for the new linear classifier that computes the class scores of our new dataset. Figure 1 depicts the overall architecture described above.
  


\section{EXPERIMENTS}
\label{sec:experiments}
To evaluate our method, we experimented on the challenging UCFSI-101 dataset and the Willow action dataset, as described below. 
\subsection{UCF-101 Sill Image (UCFSI-101) Dataset}
\label{ssec:ucf-101}

UCF-101 \cite{soomro2012ucf101} has 13320 videos from 101 action categories that spread across 5 broad groups, that is (1) Human-Object Interaction (Typing, brushing teeth, hammering, etc.), (2) Body-Motion (Baby crawling, push ups, blowing candles, walking, jumping, etc.), (3) Human-Human interaction (Head massage, salsa spin, haircut, etc.), (4) Playing Instruments (flute, guitar, piano, etc.) and (5) Sports. Considering that our method is based on the patterns associated with the human motion and also the overall shape and location of the most salient parts in the image, we get better results on Body-Motion categories as opposed to other categories that are highly dependent on detecting the presence of a specific object in the scene.

Performing the training directly on a dataset such as UCF-101 for the purpose of \textbf{\textit{video}} action classification leads to overfitting, due to the size of the dataset. However, we overcame this problem by creating a large enough annotated \textbf{\textit{still-image}} dataset by extracting and labeling over \textit{2M} video frames from the original UCF-101 dataset \cite{soomro2012ucf101}. We collected 1,585,071 frames as our training set and 617,321 frames to help form our test set. After domain-mapping of the collected still images from RGB to $I\textsubscript{POF\textsubscript{h} POF\textsubscript{v}SM }\in{\textit{POF-SM}}$, we trained our deep CNN model on this new huge dataset that we call the UCFSI-101. 

The experiments were run on a single GeForce GTX Titan GPU with 15GB of memory. To evaluate the performance of our transfer methodology, the approach mentioned in section 3 was applied on POF-SM dataset. Our method outperforms the non-pretrained baseline where CNN is trained over \textit{2M} RGB images in UCFSI-101 (i.e. the original frames extracted from the UCF-101 dataset prior to domain mapping). Our transfer method also yields more promising results compared to the case where a CNN is trained from scratch on POF-SM mapped UCFSI-101 dataset, as shown in table 1. Since we expect that CNNs learn more generic features on the bottom layers of the network, and more convoluted dataset-specific features near the top layers of the network, we considered two different scenarios for our transfer learning experiments. \begin{itemize}
\item\textit{Fine-tuning all layers:}
In this scenario, we re-trained all network parameters, including all convolutional layers on the bottom of the network.
\item \textit{Freezing the first three convolutional layers and fine-tuning the rest:} Rather than only re-training the final classifier layer from scratch, we performed fine-tuning on the last five layers.
\end{itemize} 
We further broke down our performance by 5 broad groups of classes present in the UCFSI-101 dataset. We computed the average precision of every class and then computed the mean average precision over classes in each group. In all experiments, fine-tuning the last 5 layers of the network increased the performance on all action categories. Also among all groups, we obtained the best results on the Body-Motion group, in which actions heavily depend on predicted body motion and also the most salient part of the body. Tables 1 and 2 depict the results accordingly.

\begin{table}[h]
\large
\centering
\caption{Still image action recognition results on both UCFSI-101 RGB images (used as baseline) and the POF-SM mapped UCFSI-101 dataset. Results on POF-SM UCFSI-101 dataset benefited from two different transfer learning scenarios.}
\label{table1}
\begin{tabular}{|l|l|l|ll}
\cline{1-3}
\textbf{Model}                      & \textbf{\begin{tabular}[c]{@{}l@{}}Top-1\\ Accuracy\end{tabular}} & \textbf{\begin{tabular}[c]{@{}l@{}}Top-5\\ Accuracy \footnotemark\end{tabular}} &  &  \\ \cline{1-3}
Train from scratch on RGB images    & 14.8\%                                                            & 22.1\%                                                              &  &  \\ \cline{1-3}
Train from scratch on POF-SM  & 21.2\%                                                            & 35.0\%                                                              &  &  \\ \cline{1-3}
Fine-tune all layers on POF-SM               & 41.8\%                                                            & 55.3\%                                                              &  &  \\ \cline{1-3}
Fine-tune top 5 layers on POF-SM              & \textbf{63.7\%}                                                            & \textbf{70.8\%}                                                              &  &  \\ \cline{1-3}
\end{tabular}
\end{table}
\footnotetext{5 guesses are allowed.}
\begin{table}[h]
\centering
\caption{Mean Average Precision (MAP) of our network broken down by category groups.}
\label{table2}
\large
\begin{tabular}{|l|c|c|c|c|}
\hline
\multicolumn{1}{|c|}{\textbf{Group}}                                    & \multicolumn{1}{c|}{\textbf{\begin{tabular}[c]{@{}c@{}}MAP\\ from scratch\\ on RGB\end{tabular}}} & \multicolumn{1}{c|}{\textbf{\begin{tabular}[c]{@{}c@{}}MAP\\ from scratch \\ on POF-SM\end{tabular}}} & \multicolumn{1}{c|}{\textbf{\begin{tabular}[c]{@{}c@{}}MAP \\ fine-tune\\ all layers\end{tabular}}} & \multicolumn{1}{c|}{\textbf{\begin{tabular}[c]{@{}c@{}}MAP\\ fine-tune\\ top 5 layers\end{tabular}}} \\ \hline
\begin{tabular}[c]{@{}l@{}}Human-Object\\ Interaction\end{tabular}      &                                                                                                   0.20\ & 0.17\                                                                                                        & 0.35\                                                                                                                                                                                                            & 0.56\                                                                                                                                                                                                             \\ \hline
Body-Motion                                                             &                                                                                                   0.11\ & \textbf{0.42}\                                                                                                                                                                                                              & \textbf{0.58}\                                                                                                                                                                                                            & \textbf{0.84}\                                                                                                                                                                                                             \\ \hline
\begin{tabular}[c]{@{}l@{}}Human-Human\\Interaction\end{tabular}    &                                                                                                   0.10\ & 0.35\                                                                                                                                                                                                              & 0.42\                                                                                                                                                                                                            & 0.58\                                                                                                                                                                                                             \\ \hline
\begin{tabular}[c]{@{}l@{}}Playing Musical\\ Instruments\end{tabular} &                                                                                                   0.20\ & 0.18\                                                                                                                                                                                                              & 0.30\                                                                                                                                                                                                            & 0.60\                                                                                                                                                                                                             \\ \hline
Sports                                                                  &                                                                                                   0.17\ & 0.23\                                                                                                                                                                                                              & 0.42\                                                                                                                                                                                                            & 0.65\                                                                                                                                                                                                             \\ \hline
All groups                                                              &                                                                                                   0.16\ & 0.24\                                                                                                                                                                                                              & 0.42\                                                                                                                                                                                                            & \textbf{0.65}\                                                                                                                                                                                                             \\ \hline
\end{tabular}
\end{table}
\subsection{Willow dataset}
\label{ssec:willow}

The Willow action dataset \footnote{Willow is available at: http://www.di.ens.fr/willow/research/stillactions/} contains 911 images split into seven action categories: Interacting with computer, Photographing, Playing music, Riding bike, Riding horse, Running and Walking. We used the train and test splits provided by the original authors. We also used standard data augmentation, i.e. randomly mirrored images to avoid spatial biases (such as humans always centered in the image). We then mapped all the images in the training and test sets to $I\textsubscript{POF\textsubscript{h} POF\textsubscript{v}SM }$. We removed the last fully-connected layer from our network, which was pre-trained on the generated POF-SM\textsubscript{UCFSI-101} dataset mentioned in section \ref{ssec:ucf-101}. We then treated the rest of the CNN as a fixed feature extractor for the new dataset POF-SM\textsubscript{Willow}. Moreover, we trained a linear softmax classifier for the new POF-SM\textsubscript{Willow} dataset. We further divided the 7 action categories into two main groups, Body-Motion and Non-Body-Motion actions. The first three actions, pointed out in the beginning of this section are considered as Non-Body-Motion actions since they are highly dependent on human-object interactions and the rest are considered as Body-Motion actions. We computed the Mean Average Precision (MAP) over classes in each group. We achieve the best performance on the Body-Motion group, with an MAP of 76.1\%. Table 3 shows the results on the Willow dataset.

\begin{table}[h]
\large
\centering
\caption{Comparison in (MAP) of our approach against state-of-the-art approaches on the Willow dataset.}
\label{table3}
\begin{tabular}{@{}lcclll@{}}
\cmidrule(r){1-3}
Method         & \textbf{Non-Body-Motion} & \textbf{Body-Motion} &  &  &  \\ \cmidrule(r){1-3}
Delaitre et al. \cite{delaitre2010recognizing} & 55.6                     & 62.7                 &  &  &  \\
Delaitre et al. \cite{delaitre2011learning} & 55.4                     & 70.7                 &  &  &  \\
Sharma et al. \cite{sharma2012discriminative}   & 59.0                     & 71.1                 &  &  &  \\
Sharma et al. \cite{sharma2013expanded}   & 60.1                     & 73.2                 &  &  &  \\
Khan et al. \cite{khan2014scale}     & 62.4                     & 72.2                 
&  &  &  \\
Liang et al. \cite{liang2014expressive}    & 89.0                     & 74.0                 &  &  &  \\
 \cmidrule(r){1-3}
Ours           & 62.9                        & \textbf{76.1}           &  &  &  \\ \cmidrule(r){1-3}
\end{tabular}
\end{table}

\section{CONCLUSION}
\label{sec:conclusion}

Identifying information from a single image has made major progress in recent years, including object recognition \cite{residual,spp,vgg,select-search,krizhevsky2012imagenet}, geo-localization \cite{Junejo_etal_2010,Junejo_Foroosh_2010,Junejo_Foroosh_solar2008,Junejo_Foroosh_GPS2008,junejo2006calibrating,junejo2008gps}, image classification \cite{tariq2015feature,tariq2015t}, etc. Herein, we propose a transfer learning approach for action recognition in still images using deep CNN that exploit predicted motion and saliency maps as temporal and spatial cues. The results outperform the state of the art, and are demonstrated on challenging datasets, such as the newly created UCFSI-101, which has not been previously explored in the context of still image action classification. We have shown that the proposed method yields better results by a remarkable margin compared to recent works in the literature, specifically on the actions that rely on human-body motions.

{
\bibliographystyle{plain}
\bibliography{strings,refs,foroosh}
}

\end{document}